\documentclass[10pt,twocolumn,letterpaper]{article}

\PassOptionsToPackage{table}{xcolor}

\usepackage{comment}
 \usepackage[algorithms]{wacv}    

\usepackage{newtxtext}
\usepackage{amsfonts}     
\usepackage{array}        
\usepackage{multirow}     
\usepackage{siunitx}      
\usepackage{tikz}         

\AtEndPreamble{%
  \Crefname{equation}{Equation}{Equations}%
  \crefname{equation}{Eq.}{Eqs.}%
  \Crefname{figure}{Figure}{Figures}%
  \crefname{figure}{Fig.}{Figs.}%
}

\definecolor{myblue}{HTML}{3333FF}
\definecolor{mygreen}{HTML}{D5E8D4}
\definecolor{mysalmon}{HTML}{F8CECC}
\definecolor{mybrown}{HTML}{FAD7AC}
\definecolor{myblueish}{HTML}{DAE8FC}
\definecolor{mygray}{HTML}{BAC8D3}
\definecolor{mypurple}{HTML}{D0CEE2}
\definecolor{myyellow}{HTML}{FFF2CC}

\newcommand{\PLACEHOLDER}[1]{%
  \ifmmode \textcolor{red}{[#1]}\else \textcolor{red}{[#1]}\fi}

\newcommand{\specialcell}[2][l]{%
  \begin{tabular}[#1]{@{}l@{}}#2\end{tabular}}

\newcommand{\cube}[1]{
    \begin{tikzpicture}[scale=#1]
        \draw[fill=myyellow] (0,0) rectangle (1,1); 
        \draw[fill=myyellow] (0.3,0.3) rectangle (1.3,1.3); 
        \draw[fill=myyellow] (0,0) -- (0.3,0.3) -- (1.3,0.3) -- (1,0) -- cycle; 
        \draw[fill=myyellow] (1,1) -- (1.3,1.3) -- (1.3,0.3) -- (1,0) -- cycle; 
        \draw[fill=myyellow] (0,1) -- (0.3,1.3) -- (1.3,1.3) -- (1,1) -- cycle; 
    \end{tikzpicture}
}

\newcommand{\Oshape}[1]{
    \begin{tikzpicture}[scale=#1]
        \draw (0,0) circle (1.5cm); 
        \foreach \a in {0,30,...,330} {
          \fill (\a:1.5cm) circle (10pt); 
        }
    \end{tikzpicture}
}

\definecolor{wacvblue}{rgb}{0.21,0.49,0.74}
\usepackage[pagebackref,breaklinks,colorlinks,allcolors=wacvblue]{hyperref}

\def\wacvPaperID{258} 
\def\confName{WACV}
\def\confYear{2027}

\title{RECALL: Real-Time Recalibration of Multi-Camera Rigs via Learned OEM-Anchored Residuals}

\author{Mario Alberto Tapia de la Cruz \quad Charalambos Poullis\\
Immersive and Creative Technologies (ICT) Lab, Concordia University\\
Montreal, Canada\\
{\tt\small mario.tapiadelacruz@mail.concordia.ca \quad charalambos@poullis.org}
}

\begin{document}
\maketitle
\begin{abstract}
Multi-camera rigs ship pre-calibrated from the manufacturer, but their calibration goes stale each time the rig is repositioned, and drifts continuously through wear, thermal cycling, and vibration. Recovering it on-the-fly is hard: classical bundle adjustment needs a fresh multi-view capture and a global optimization whose cost grows with the rig, ruling out real-time use. In this paper we present \textsc{RECALL}, a feed-forward method that recalibrates a \emph{movable} multi-camera rig in a single pass. From the nominal OEM calibration as an anchor and a brief observation of a known object, it predicts each camera's current intrinsics and extrinsics, written as a shared \emph{global rotation} of the rig composed with a small per-camera \emph{drift}. Instead of regressing the calibration in one shot, \textsc{RECALL} refines its estimate with a learned, rig-coupled recurrent operator that reprojects through a differentiable projection and applies attention across the cameras at each step, and \emph{solves} for the shared rotational gauge in-network with a learned Gauss-Newton step rather than regressing it. Because the ground truth is synthesizable, we train entirely on data generated on-the-fly and need no captured dataset. On a synthetic 10-camera benchmark, \textsc{RECALL} is more accurate than per-camera and rig-coupled joint bundle adjustment ($0.26^\circ$ vs $0.47^\circ$ rotation, $5.4$ vs $8.9$\,px focal) and runs about $10^5\times$ faster at ${\sim}0.12$\,ms per rig. On real data (TotalCapture, CMU Panoptic) it \emph{matches or beats} offline rig-coupled bundle adjustment on absolute and relative rotation while running four-to-five orders of magnitude faster, and on the Panoptic surround rig it also matches the strongest image-only foundation models (VGGT, VGGT-$\Omega$, $\pi^3$) on relative pose while staying metric and about $15\times$ more accurate in focal length. Code: \url{https://github.com/theICTlab/RECALL-multicam}
\end{abstract}
\vspace{-20pt}
\section{Introduction}
\label{sec:intro}

Camera calibration, the estimation of a camera's intrinsic and extrinsic parameters, is fundamental to computer vision and especially critical in high-precision settings such as robotics, autonomous platforms, AR/VR, and motion-capture and broadcast studios. Traditional methods capture a known object from many viewpoints, extract features and correspondences, and run a global optimization (typically bundle adjustment) whose cost grows rapidly with the number of cameras, images, and correspondences. This makes them unsuitable for real-time use.

Commercial multi-camera systems ship pre-calibrated, but this factory (OEM) calibration drifts over time due to: thermal expansion and contraction as the system heats and cools, mechanical shock and vibration from transport, handling, or re-mounting, gradual wear of the mounts and fixtures, and dust, smudges, or debris on the lenses and fiducials. Each of these shifts a camera's intrinsics or extrinsics slightly, degrading downstream measurements such as 3D tracking. 
Re-calibrating in the field is impractical: Coordinate Measuring Machines are costly and do not capture in-field drift, classical bundle adjustment needs a fresh multi-view capture and a slow optimization, and epipolar error-detection only flags a problem without correcting it in real time. Fast in-field recalibration is therefore needed to keep deployed rigs accurate.

We target \emph{movable} multi-camera rigs that ship pre-calibrated and are repositioned in deployment: their cameras are rigidly mounted relative to one another by the manufacturer, but the whole rig may undergo a bounded global reorientation at each setup. Recovering the \emph{current} calibration thus has two coupled parts: the rig's \emph{global rotation} relative to the working volume, and the small \emph{per-camera drift} of intrinsics and extrinsics from their nominal (OEM) values. We recover both in real time from a brief observation of a known object, without the offline multi-view capture and optimization that conventional calibration requires. We assume this global reorientation is \emph{bounded} (tilted within an operating range, not spun freely about the optical axis), which makes a direct feed-forward prediction well posed.

We present \textsc{RECALL} (REal-time multi-CAmera recaLibration), which treats the OEM calibration as an \emph{anchor} and, from a brief observation of a known object, predicts each camera's current calibration in a single feed-forward pass. We parameterize the prediction as a shared \emph{global rotation} of the rig composed with a small OEM-anchored per-camera \emph{drift}. Instead of regressing the calibration in one shot, \textsc{RECALL} refines it with a learned, \emph{rig-coupled recurrent} operator. At each step, the operator reprojects its current estimate and applies attention accross the cameras. Crucially, it \emph{solves} for the shared rotational gauge in-network with a learned Gauss-Newton step rather than regressing it. Casting calibration as learning is well established~\cite{bogdan2018deepcalib,kendall2015posenet,hold2018perceptual}, and our recurrent reprojection-feedback design builds on RAFT and DROID-SLAM~\cite{teed2020raft,teed2021droid}; our contribution is the deployment-oriented formulation that recovers the shared global rotation and full per-camera calibration jointly, anchored to the OEM prior, with a learned gauge solver that lets a single pass match or beat the bundle adjustment it replaces at run time. Our main contributions are:
\begin{itemize}
    \item A \emph{feed-forward formulation for movable-rig recalibration} that anchors to the OEM calibration and jointly recovers a shared global rig rotation and small per-camera intrinsic and extrinsic drift in a single pass, trained entirely on synthesizable on-the-fly data with no captured calibration dataset.
    \item A \emph{rig-coupled recurrent refinement operator} that reprojects the current estimate at each step, applies cross-camera attention to exploit rig rigidity and multi-view consistency, and jointly refines extrinsics, intrinsics, and distortion, reducing parameter error over its iterations.
    \item A \emph{learned Gauss-Newton gauge solver}:  rather than regress the ill-conditioned shared rotational gauge, \textsc{RECALL} \emph{solves} for it in-network with a damped, robustly-weighted Gauss-Newton step and network-predicted confidence and damping. This closes the absolute-orientation gap to per-instance bundle adjustment, so a single pass matches or beats offline joint BA on rotation.
\end{itemize}

\section{Related Work}
\label{sec:related_work}
Traditional calibration, including the Radial Distortion Model~\cite{duane1971close}, DLT~\cite{abdeldirectlineartransformation}, Tsai's~\cite{tsai1987versatile}, Zhang's~\cite{zhang2000flexible} methods, and fish-eye lens calibration~\cite{shah1994simple}, relies on handcrafted features and static multi-view capture, which is ill-suited to real-time or multi-camera use. Perspective-$n$-point solvers~\cite{nakano2016versatile,urban2016mlpnp,haner2015absolute} improve pose estimation but still face real-time challenges.

Neural networks have shifted camera calibration research~\cite{liao2023deep}. Methods such as CalibBD~\cite{nguyen2022calibbd}, a CNN-based extrinsic calibration approach~\cite{itu2017automatic}, and PoseNet~\cite{kendall2015posenet} target extrinsic calibration, while DeepFocal~\cite{workman2015deepfocal}, DeepCalib~\cite{bogdan2018deepcalib}, and other learned approaches~\cite{cramariuc2020learning,hagemann2023deep} target intrinsics, using deep learning for more adaptable and potentially real-time calibration. More recent single-image methods predict perspective or geometric fields and refine them with geometric optimization~\cite{jin2023perspective,veicht2024geocalib}. These networks typically require extensive training data, target a single camera, and do not exploit a known nominal calibration.

For multi-camera setups, self-calibration methods~\cite{svoboda2005convenient,heikkila1997four,faugeras1992camera,fraser1997digital,hartley1994self} enable 3D reconstruction and multi-view analysis but are challenged by real-time, dynamic operation, while learned joint intrinsic and extrinsic estimation~\cite{hold2018perceptual,butt2022camera,lopez2019deep,wakai2021deep} still trades off efficiency against accuracy.

\noindent\textbf{Feed-forward geometric foundation models.} A recent line predicts scene geometry and cameras directly from images in one pass. DUSt3R~\cite{wang2024dust3r} and MASt3R~\cite{leroy2024mast3r} regress per-pixel pointmaps, and VGGT~\cite{wang2025vggt}, VGGT-$\Omega$~\cite{vggtomega2025}, and $\pi^3$~\cite{pi3_2025} jointly predict cameras, depth, and geometry for a set of views. These models are prior-free and recover cameras only up to a similarity (scale and global pose), so they are not metric and do not exploit a known nominal calibration. We compare against them on real data (\cref{subsec:realdata}), lifting their predictions to metric with the OEM prior for fairness. As the closest learned alternative to recalibration, they are strong on relative pose but, on wide-baseline surround rigs, far less accurate in intrinsics and orders of magnitude slower than a prior-anchored predictor. Online or targetless extrinsic recalibration and SLAM-coupled self-calibration~\cite{teed2021droid} refine calibration from motion, but optimize per-sequence rather than amortizing a constant-time predictor over a fixed rig.

\noindent\textbf{Learned and differentiable geometric optimization.} A separate line embeds the geometric solver in the network: differentiable nonlinear-least-squares layers~\cite{pineda2022theseus,wang2023pypose,gould2021deep} and learned dense bundle adjustment (BA-Net~\cite{tang2019banet}, learned Levenberg-Marquardt damping; DROID-SLAM~\cite{teed2021droid}, a Gauss-Newton solve with learned confidence), while end-to-end differentiable PnP layers~\cite{chen2020bpnp,chen2022epropnp} back-propagate through single-camera pose with fixed intrinsics. The differentiable solver itself is not new: learned damping and confidence weights and the recurrent reprojection-feedback design (RAFT~\cite{teed2020raft}, DROID-SLAM) are established, and cross-camera coupling is what bundle adjustment already does~\cite{triggs2000bundle}. These solvers all optimize \emph{per instance at inference}. \textsc{RECALL} differs in three respects that matter for rig recalibration: it \emph{amortizes} the optimization to training and runs one feed-forward pass over a known rig at deployment; it is \emph{anchored to the OEM prior}, which keeps per-camera drift small and regressable; and it \emph{restricts} the in-network Gauss-Newton step to the single ill-conditioned shared gauge (the rig's global rotation, a $3{\times}3$ solve, \cref{eq:gngauge}), which is unobservable per camera and emerges only from cross-camera coupling~\cite{triggs2000bundle}, while regressing the well-anchored remainder.

Unlike the methods above, our method runs in real time and predicts camera parameters under drift of the OEM intrinsic and extrinsic values. A streamlined on-the-fly synthesis procedure lets it generalize to arbitrary multi-camera configurations and arbitrary calibration objects.

\section{Camera Model}
\label{sec:camera_model}
\textsc{RECALL} recovers the parameters of the standard pinhole camera model with lens distortion, whose image-formation map is the differentiable component used to express the reprojection term of our loss (\cref{sec:methodology}). A world point $\mathbf{P}=[X,Y,Z]^T$ with camera orientation $\mathbf{R}$ and camera center $\mathbf{t}$ maps to camera coordinates $\mathbf{P}^{C}=\mathbf{R}^\mathbf{{\top}}(\mathbf{P}-\mathbf{t})$ and projects to the image as
\begin{equation}
  \scalebox{0.9}{$
    \left[\begin{smallmatrix} x \\ y \\ 1 \end{smallmatrix}\right]
    = \frac{1}{Z^{C}}
    \underbrace{\left[\begin{smallmatrix} f_x & 0 & c_x \\ 0 & f_y & c_y \\ 0 & 0 & 1 \end{smallmatrix}\right]}_{\mathbf{K}}
    \mathbf{P}^{C}
  $}
  \label{eq:projection}
\end{equation}
with focal lengths $f_x,f_y$ and principal point $(c_x,c_y)$. We model lens distortion with the standard $5$-coefficient OpenCV model, with three radial ($k_1,k_2,k_3$) and two tangential ($p_1,p_2$) terms applied to the normalized coordinates before the intrinsic scaling. Rotations use the continuous $6$D parameterization of Zhou \etal~\cite{Zhou_2019_CVPR} (the first two columns of $R$, recovered by Gram-Schmidt), which is discontinuity-free and guarantees a valid map to $SO(3)$, unlike quaternions or Euler angles. The full per-camera state is thus the $15$ parameters $[\,\mathbf{R}\,|\,\mathbf{t}\,]$, $(f_x,f_y)$, $(c_x,c_y)$, $(k_1,k_2,k_3,p_1,p_2)$ (the network emits $18$ numbers per camera, as the continuous $6$D rotation uses $6$ values for the $3$ rotational DoF); the map is fully differentiable, supplying the reprojection residual that drives each refinement step.

\section{Methodology}
\label{sec:methodology}
\begin{figure*}[!ht]
    \centering
    \includegraphics[width=0.70\textwidth]{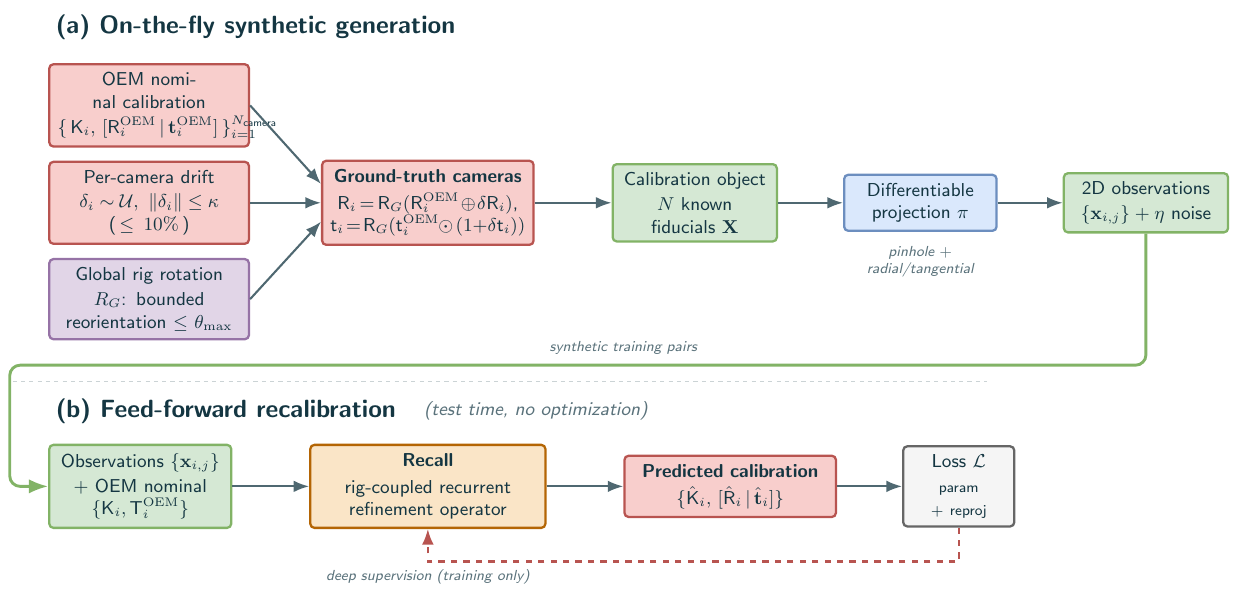}
    \caption{\textbf{Method overview.} \textbf{(a)}~On-the-fly synthesis: from the OEM calibration we sample a bounded global rotation $R_G$ (global reorientation $\le\theta_{\max}$) and small OEM-anchored per-camera drift ($\le\kappa$), compose ground-truth cameras via \cref{eq:decomp}, and project the known object to noisy $2$D observations, regenerated every iteration with no captured dataset. \textbf{(b)}~At test time the observations and OEM anchor pass once through \textsc{RECALL}, the rig-coupled recurrent operator (\cref{fig:architecture}), which predicts the full per-camera calibration; during training the differentiable projection supplies the reprojection term, reaching every parameter head.}
    \label{fig:technical_overview}
\end{figure*}

Given a \emph{movable} multi-camera rig whose nominal (OEM/factory) calibration is known, our goal is to recover the rig's \emph{current} calibration in a single feed-forward pass, in real time, from a brief observation of a known calibration object. The current calibration differs from nominal for two reasons: the rig has undergone a bounded global reorientation, and each camera has \emph{drifted} slightly from its factory values through wear, thermal cycling, and vibration. We model these jointly. For camera $i,\ 0 \le i < N_\text{camera}$, with nominal extrinsics $T_i^{\mathrm{OEM}}=[R_i^{\mathrm{OEM}}\,|\,t_i^{\mathrm{OEM}}]$ and intrinsics $\gamma_i^{\mathrm{OEM}} = (f_i^{\mathrm{OEM}},p_i^{\mathrm{OEM}},k_i^{\mathrm{OEM}})$, where $f_i=(f_{x,i},f_{y,i})$, $p_i=(c_{x,i},c_{y,i})$, and $k_i$ denotes distortion coefficients, 
the current extrinsics are modeled as
\begin{equation}
\label{eq:decomp}
\begin{aligned} 
R_i &= R_G\left(R_i^{\mathrm{OEM}}\oplus \delta R_i\right), \\ 
t_i &= R_G\left(t_i^{\mathrm{OEM}}\odot(1+\delta t_i)\right),
\end{aligned} 
\end{equation}
where $R_G\in SO(3)$ is a \emph{shared global rotation} of the whole rig (a bounded global reorientation common to all cameras), and $\delta_i^{\mathrm{ext}} = (\delta R_i, \delta t_i)$.
The current intrinsics are modeled as  $\gamma_i=\gamma_i^{\mathrm{OEM}}\odot(1+\delta_i^{\mathrm{int}})$, where $\delta_i = (\delta_i^{\mathrm{ext}}, \delta_i^{\mathrm{int}})$ denotes the per-camera extrinsic and intrinsic drift. Here $\oplus$ composes a small relative-rotation drift onto an orientation, $R^{\mathrm{OEM}}\oplus\delta R := \delta R\cdot R^{\mathrm{OEM}}$, and $\odot$ denotes the element-wise multiplicative drift that scales translation and intrinsics about their nominal values, $v^{\mathrm{OEM}}\odot(1+\delta) := v^{\mathrm{OEM}}\cdot(1+\delta)$. The network is conditioned on the OEM calibration and on a $2$D observation of the calibration object, and predicts, for every camera jointly, the absolute current calibration $\hat{C}_i$ anchored at $C_i^{\mathrm{OEM}}$. This formulation is amortized: all cost is paid once at training time, and recalibration at deployment reduces to one network evaluation. Anchoring to the OEM prior keeps the per-camera drift small and well posed; the shared global rotation $R_G$ captures the (potentially large) common rig reorientation that a purely per-camera residual could not represent. \Cref{fig:technical_overview} provides an overview.

\subsection{Movable-Rig Training Data Synthesis}
\label{subsec:train_data_generation}
A defining property of the rigs we target (AR/VR headsets, robot and vehicle sensor suites, motion-capture and broadcast arrays) is that the cameras are rigidly mounted relative to one another by the OEM, while the \emph{whole rig} is movable and may undergo a bounded global reorientation in deployment. Recalibration therefore has two parts: estimating the rig's bounded \emph{global rotation} $R_G$ and the small per-camera \emph{drift} from nominal (\cref{eq:decomp}). A key advantage of this task is that ground truth is \emph{synthesizable}: given the OEM calibration we can sample a global rotation and per-camera drift, project the known object, and obtain a perfectly labeled training pair, so we generate data \emph{on-the-fly} every iteration and never capture a calibration dataset. Each sample draws a global rotation and per-camera drift, composes them into ground-truth cameras via \cref{eq:decomp}, and projects a known object at a fixed location (optionally over $M$ short snapshots). The rig's reorientation, not the object, supplies the viewpoint variation.

\noindent\textbf{Bounded global rotation.} We sample the shared global rotation $R_G$ as a rotation by an angle $\alpha \sim \mathcal{U}(0, \theta_{\max})$ about a uniformly random axis (with $\theta_{\max}=30^\circ$ in our experiments). This models a boom-mounted rig whose orientation changes within a bounded operating range and is not freely rotated about the optical axis. Unlike a pure per-camera residual, the explicit $R_G$ lets the model represent a large common reorientation while keeping each camera's remaining drift small.

\noindent\textbf{Bounded per-camera drift.} Given $R_G$, each camera receives an independent drift whose magnitude is scaled by $\kappa$ (we use $\kappa\le 0.10$): a small relative rotation and translation on its extrinsics and a scaled drift of its focal length, principal point, and distortion. The rotation drift is a bounded relative angle, and the translation/intrinsic drift is a \emph{two-sided} perturbation about the nominal value; concretely, at $\kappa{=}0.10$ each camera deviates by a rotation of up to ${\sim}2.9^\circ$ and a translation/intrinsic jitter of at most ${\pm}5\%$ of nominal (since the perturbation is symmetric about nominal).  Each rig first draws a drift level uniformly in [0, $\kappa$], shared by its cameras, within which each camera draws its drift independently. Because the drift is independent across cameras and re-drawn every iteration, the network sees a continuum of drift states, 
letting a single model recalibrate the rig across its operating envelope. 

\noindent\textbf{Differentiable projection.} The known 3D fiducials are projected with the differentiable, vectorized pinhole-plus-distortion model of \cref{sec:camera_model} (\cref{eq:projection}), so the gradient of any image-space loss reaches all parameter heads simultaneously; we use it both to synthesize observations and to define the reprojection loss. The framework supports arbitrary rig topologies (T-, U-, and O-shaped arrays) and calibration objects (cube, sphere), with results across both reported in the supplementary.

\subsection{Rig-Coupled Recurrent Refinement}
\label{sec:model}

\begin{figure*}[t]
    \centering
    \includegraphics[width=0.70\textwidth]{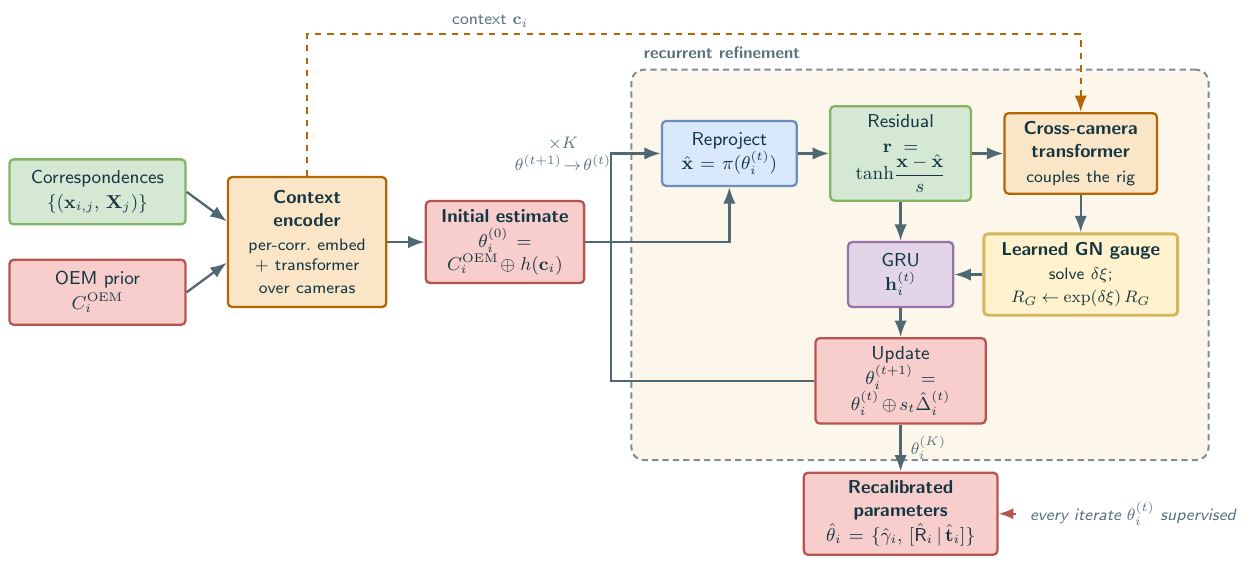}
    \caption{\textbf{The \textsc{RECALL} operator.} A context encoder embeds each camera's 2D-3D correspondences and OEM calibration, applying attention \emph{across the cameras} to a context $\mathbf{c}_i$ from which OEM-anchored heads predict an initial estimate $\theta_i^{(0)}$. It is refined for $K$ steps: reproject through the differentiable projection, form a bounded residual, applying attention across the cameras (rig coupling), update a per-camera GRU, and apply a bounded update; in parallel a learned Gauss-Newton step \emph{solves} for the shared rotational gauge $R_G$ (\cref{eq:gngauge}) from the same residual. Every iterate is supervised; the final one is the output.}
    \label{fig:architecture}
\end{figure*}
Because one-shot regression must approximate the nonlinear full calibration problem of a movable rig, \textsc{RECALL} predicts an initial estimate and then \emph{refines it with geometry in the loop}.
Since cameras of a rig are coupled through the shared body and a commonly-observed object, we make the refinement \emph{rig-coupled} (\cref{fig:architecture}). The rig observes the known object (optionally at $M$ short snapshots), yielding per-camera 2D-3D correspondences.

\noindent\textbf{Context encoder and initialization.} We encode the rig once. Each camera contributes a token that flattens its 2D-3D correspondences (preserving point identity) together with the OEM calibration; a transformer over the cameras produces a context $\mathbf{c}_i$ reflecting the shared rig and common object. From $\mathbf{c}_i$, six OEM-anchored heads predict an initial calibration estimate for camera $i$, $\theta_i^{(0)} = C_i^{\mathrm{OEM}} \oplus h(\mathbf{c}_i)$, with the final layer of each head initialized to small weights so the model starts near the prior and the encoder receives gradient from the first step. The initial estimate already accounts for the shared global rotation; the refinement then tightens it. Rotations use the continuous 6D parameterization of Zhou \etal~\cite{Zhou_2019_CVPR}, mapped to $SO(3)$ before projection. 

\noindent\textbf{Coupled update operator.} The estimate is refined for $K$ steps. At step $t$ we reproject the current estimate through the differentiable projection (\cref{eq:projection}) and form a bounded per-fiducial reprojection residual $r_{i, j}^{(t)} = \tanh\!\big((\mathrm{x}_{i, j}^{\text{obs}} - \pi(\mathrm{x}_j;\theta_i^{(t)}))/s\big)$ with a stop-gradient on the reprojected measurement, which keeps the recurrence stable; in the first two of the $K$ steps, $\operatorname{asinh}$ replaces $\tanh$ without the stop-gradient. The per-camera residual features are passed through a transformer that applies attention \emph{across the cameras} before a gated recurrent unit updates each camera's hidden state and emits a bounded parameter update $\theta_i^{(t+1)} = \theta_i^{(t)} \oplus s_t\,\hat{\Delta}_i^{(t)},\qquad t=0,\dots,K-1$, with the final iterate $\hat{C}_i = \theta_i^{(K)}$. The cross-camera attention \emph{inside} the loop lets each camera's update respect rig rigidity and the multi-view consistency of the common object. We supervise all iterates with exponentially increasing weight (\cref{sec:training}); the refinement reduces both parameter and reprojection error monotonically over its steps (supplementary).

\noindent\textbf{Solving for the shared rotational gauge.} We \emph{solve} for the shared global \emph{rotation} $R_G$ inside the loop, rather than regress it. The per-camera drift $\delta_i$ is small and OEM anchored, but the shared global rotation $R_G$ is the ill-conditioned part of \cref{eq:decomp}:  it couples all cameras, 
so a pooled \emph{regressor} for $R_G$ leaves an \emph{absolute} orientation residual even when the inter-camera (relative) geometry is accurate, the very gauge that bundle adjustment fixes by joint optimization~\cite{triggs2000bundle}. Once per refinement step, we instead take a single Gauss-Newton iteration restricted to a global-rotation increment $\delta\xi\in\mathbb{R}^3$ (an axis-angle vector) shared by all cameras. With $r$ the reprojection residual stacked over all in-view fiducials, $J=\partial r/\partial\xi|_{\xi=0}\in\mathbb{R}^{|r|\times 3}$ its Jacobian, computed by forward-mode automatic differentiation in one pass through the differentiable projection; $W$ a diagonal per-measurement confidence matrix, with each camera's residuals weighted by a confidence $w_i{>}0$ from a linear head on that camera's in-loop features; and $\lambda\in\mathbb{R}^3{>0}$ a per-axis damping from a small MLP reading the rig-pooled features (both softplus-activated); the increment and update are
\begin{equation}
\label{eq:gngauge}
\begin{aligned}
\big(J^\top WJ+\operatorname{diag}(\lambda)\big)\delta\xi= -J^\top Wr, \quad R_G \leftarrow \exp(\delta\xi) R_G .
\end{aligned}
\end{equation}
a $3{\times}3$ solve that is batched over the rig and fully differentiable, so $W$ and $\lambda$ are learned end-to-end against the supervised loss. Because the gauge is \emph{computed} from the measurement rather than guessed, the single feed-forward pass already attains a global orientation of per-instance-optimization quality. This specializes learned dense bundle adjustment~\cite{tang2019banet,teed2021droid}, which optimizes poses and depth per sequence, to an OEM-anchored, gauge-restricted Gauss-Newton step for a movable rig (\cref{sec:related_work}).

\noindent Overall, \textsc{RECALL} refines the full OEM-anchored calibration (intrinsics, extrinsics, and distortion) while solving the shared rotational gauge in-network, yielding a deployment-oriented, real-time recalibration formulation.

\subsection{Loss}
\label{subsec:loss}
The training objective compares the recalibrated prediction $\hat{C}_i$ against the ground-truth current calibration $C_i$ used to synthesize the sample, using both parameter-space and image-space supervision terms.

\noindent\textbf{Parameter-space term.} $\mathcal{L}_{\mathrm{diff}}$ compares predicted and ground-truth parameters. Focal length and principal point are each normalized by the mean deviation between the OEM nominal and the ground truth over the batch. Translation is normalized by its own magnitude instead. Distortion is supervised only through the reprojection term. Orientation is supervised with a \emph{chordal} rotation loss $\mathcal{L}_{R} = \lVert \hat{R} - R \rVert_F^2$, a smooth surrogate for the geodesic angle. Unlike $\arccos$, it has no gradient singularity near zero error, which matters for stable convergence to small residuals.

\noindent\textbf{Observability-gated focal.} A single snapshot ($M{=}1$) only weakly constrains focal length, which trades against depth, so a fixed-weight focal loss either overfits observation noise or is swamped by better-observed parameters. We supervise focal with a heteroscedastic negative log-likelihood~\cite{kendall2017uncertainties}: a per-camera head predicts a log-variance $\beta_i$, and the focal term is $\mathcal{L}_{\mathrm{focal}}
= \tfrac12 e^{-\beta_i}\,(\log\hat{f}_i - \log f_i)^2 + \tfrac12 \beta_i$. 
The learned variance \emph{gates} the signal: it rises where focal is unobservable, shrinking its gradient and deferring to the OEM prior, and falls when additional snapshots make focal observable, avoiding fitting a quantity the geometry does not determine.

\noindent\textbf{Image-space term.} $\mathcal{L}_{\epsilon}$ is the reprojection RMSE between the 2D fiducial projections obtained with the predicted and the ground-truth parameters, computed through the differentiable projection. We treat reprojection as a \emph{secondary} signal: a single per-camera view is consistency-limited, since reprojection can be driven low by trading focal length against depth, so an independent per-camera optimizer reaches near-zero reprojection with incorrect parameters (\cref{sec:experiments}). The parameter-space terms are therefore the primary supervision, with reprojection acting as a geometric-consistency regularizer.

Two further terms supervise the shared gauge: $\mathcal{L}_{G}=\lVert \hat{R}_G - R_G\rVert_F^2$ chordally supervises the in-loop \emph{solved} global rotation against ground truth (training the Gauss-Newton step's learned $W$ and $\lambda$ end-to-end), and a gauge-pin $\mathcal{L}_{\mathrm{pin}}=\lVert \tfrac{1}{N_\text{camera}}\sum_i \delta R_i - I\rVert_F^2$ keeps the mean per-camera rotational drift at identity, so absolute orientation is carried by $R_G$ rather than absorbed into the per-camera drift. The total loss is their weighted sum; beyond the gauge-pin we add no distortion-magnitude, focal-equality, or principal-point-centering penalties.

\subsection{Training}
\label{sec:training}
We train with a single-phase objective
$\mathcal L =
\sum_{t=0}^{K} \lambda_{\kappa}
(\lambda_R L_R^{(t)} + \lambda_tL_t^{(t)} + L_f^{(t)} + L_p^{(t)} + \lambda_\epsilon L_\epsilon^{(t)})
+ \lambda_G L_G + \lambda_{\mathrm{pin}}L_{\mathrm{pin}}$.
Thus later, more-refined iterates are weighted most. Unless stated otherwise, we use $\lambda_{\kappa} = 0.8^{K-t}$, $\lambda_R=100$, $\lambda_t =  3$, $\lambda_\epsilon=0.02$, $\lambda_G=30$, and $\lambda_{\mathrm{pin}}=10$, with the remaining parameter terms drift-normalized, and train for $6000$ iterations at batch size $256$. A bounded global rotation and per-camera drift are re-sampled every iteration (\cref{subsec:train_data_generation}), exposing the model to a continuum of reorientations and drifts such as those caused by repositioning, thermal expansion, and mechanical shock. We optimize with Adam under a warmup-then-cosine learning-rate schedule and clip gradients for stability.

\vspace{-5pt}
\section{Experimental Results}
\label{sec:experiments}
We evaluate \textsc{RECALL} as an \emph{online recalibration} method: from a movable rig's nominal (OEM/factory) calibration and a brief observation of a known object, it predicts each camera's current intrinsics and extrinsics (a shared global rotation and OEM-anchored per-camera drift) in one feed-forward pass. The synthetic results (\cref{tab:main_results}) are from real runs and labeled synthetic throughout; \emph{we do not report any synthetic number as real.} We then validate on two calibrated benchmarks (TotalCapture and CMU Panoptic, \cref{subsec:realdata}): the rigs' published calibrations and captured 3D structure are real, while the drift and global reorientation are injected under controlled magnitudes, and \textsc{RECALL}, trained on synthetic (Panoptic) or training-subject (TotalCapture) structures, is evaluated on the real captured structure and compared on the Panoptic surround rig against image-only geometric foundation models (VGGT, VGGT-$\Omega$, $\pi^3$).

\subsection{Synthetic Benchmark}
\label{subsec:synth_protocol}
For each sampled rig, the fixed object is reprojected with the differentiable model of \cref{sec:camera_model}; detections are perturbed by independent Gaussian noise ($0.5$\,px). \textsc{RECALL} receives the $2$D observations and OEM anchor and predicts the full per-camera calibration; supervision uses the losses of \cref{subsec:loss}, summed over all refinement iterates.

We instantiate the benchmark primarily on an O-shape ($N_\text{camera}=10$ \Oshape{0.08}) rig with a cube ($N_{fid.}=8$ \cube{0.15}) object, over the operating range $\kappa\le0.10$ and $\theta_{\max}=30^\circ$. Each cell is the mean over $1000$ test rigs for Stale-OEM and \textsc{RECALL} and over the first $20$ of them for the slower per-rig baselines, on a test set held fixed across methods; variance over three training seeds is reported in the suppl. 

\subsection{Baselines}
\label{subsec:baselines}
We compare against a do-nothing lower bound and two BA baselines, all initialized from the OEM prior. \textbf{Stale-OEM} uses the factory calibration unchanged ($\hat{C}_i{=}C_i^{OEM}$), quantifying the error of ignoring both reorientation and drift. \textbf{Per-camera BA} initializes each camera independently with EPnP and refines it with Levenberg-Marquardt over its full $15$-parameter calibration, ignoring rig rigidity and the shared object. We run it to convergence (the fraction reaching the iteration cap is negligible). \textbf{Rig-coupled joint BA} is the strong, structurally-matched opponent: one Levenberg-Marquardt solve over a shared global rotation $R_G$ and all cameras' OEM-anchored drift, initialized from per-camera PnP decomposed into $R_G$ and drift, exploiting the same rig structure as \textsc{RECALL} (an optional OEM-prior regularizer changes results negligibly). Learned dense-BA and SLAM methods such as DROID-SLAM~\cite{teed2021droid} target a per-sequence video SLAM from scratch, without an OEM anchor, and cannot be run directly on this prior-anchored recalibration task; the rig-coupled joint BA, which performs the same joint optimization classically, is therefore the closest directly-comparable opponent and our primary baseline. We further compare against two bounds-aware alternatives that avoid a full joint solve. \textbf{Method A} initializes each camera's pose from the OEM intrinsics via a robust PnP solver (RANSAC+EPnP, or REPPnP~\cite{Ferraz_2014_CVPR}), averages the shared global rotation across cameras with $L_1$ Weiszfeld rotation averaging~\cite{Hartley_2011_CVPR}, and truncates each camera's pose drift to RECALL's own operating bounds. \textbf{Method B} extends Method A with bound-constrained, iterative per-camera intrinsics refinement on the fixed poses, freezing each parameter once it reaches its bound. 
We report accuracy and latency for all, since per-frame optimization is the natural classical alternative for online use.

\subsection{Metrics}
\label{subsec:metrics}
\vspace{-5pt}
Because a single per-camera view is consistency-limited (an optimizer can trade focal against depth to near-zero reprojection with wrong parameters), low reprojection does not imply correct parameters; we report parameter-space errors as primary, reprojection as diagnostic. Per-camera metrics are absolute rotation (geodesic to GT, including $R_G$); relative rotation (inter-camera angle vs.\ camera $0$, invariant to $R_G$); translation (center distance); focal and principal point (px); distortion (mean absolute error over the coefficients that genuinely vary for a given rig, excluding those pinned at their nominal value) and latency (\cref{subsec:latency}). On real data we add the gauge-invariant RRA@$5^\circ$/RTA@$5^\circ$/AUC@$30^\circ$ (\cref{subsec:realdata}).

\begin{table*}
\caption{\textbf{Synthetic movable-rig recalibration} (all entries \textbf{synthetic}). Config: $N_\text{camera}{=}10$ (O-shape), $N_{fid.}{=}8$ (cube), drift $\kappa{=}0.1$, tilt $30^\circ$, $0.5$\,px noise; mean over $3$ seeds (std in supp.); latency on an NVIDIA H100. Among recovering methods (stale row excluded), \textbf{best} per metric in bold, \underline{second best} underlined; all metrics lower-is-better ($\downarrow$).}
\centering
    \resizebox{0.81\textwidth}{!}{%
      \rowcolors{2}{white}{gray!25}
      \begin{tabular}{@{}l|c|c|c|c|c|c|c@{}}
        \toprule
        \textbf{Method} &
        \specialcell[t]{\textbf{Abs. rot}~$\downarrow$\\(\si{\degree})} &
        \specialcell[t]{\textbf{Rel. rot}~$\downarrow$\\(\si{\degree})} &
        \specialcell[t]{\textbf{Translation}~$\downarrow$\\(units)} &
        \specialcell[t]{\textbf{Focal}~$\downarrow$\\(px)} &
        \specialcell[t]{\textbf{Princ. pt.}~$\downarrow$\\(px)} &
        \specialcell[t]{\textbf{Dist. err}~$\downarrow$} &
        \specialcell[t]{\textbf{Latency}~$\downarrow$\\(ms)} \\
        \midrule
        Stale-OEM (do-nothing)          & 15.34 & 1.10 & 42.03 & 20.64 & 9.98  & 0.00019   & $\approx 0$ \\
        Per-camera BA                   & 2.30 & 4.72 & 8.41  & 503.88 & 39.43 & 751.49030 & 682.21 \\
        Rig-coupled joint BA            & \underline{0.47} & \underline{0.61} & \underline{1.30}  & \underline{8.88}   & \underline{8.14}  & 0.89187   & 16767 \\
        Method A (RANSAC+EPnP)              & 0.65 & 1.00 & 1.94  & 17.03  & 8.21  & \textbf{0.00016} & \underline{5.26} \\
        Method B (RANSAC+EPnP)               & 0.65 & 1.00 & 1.94  & 12.91  & 9.98  & \underline{0.00075} & 61.81 \\
        Method A (RePPnP)          & 0.59 & 0.83 & 1.64  & 17.03  & 8.21  & \textbf{0.00016} & 10.48 \\
        Method B (RePPnP)          & 0.59 & 0.83 & 1.64  & 11.02  & 9.51  & \underline{0.00075} & 65.97 \\
        \textsc{RECALL} (ours)          & \textbf{0.26} & \textbf{0.37} & \textbf{0.68} & \textbf{5.43} & \textbf{4.65} & 0.00123   & \textbf{0.12} \\
        \bottomrule
      \end{tabular}
      }
\label{tab:main_results}
\vspace{-10pt}
\end{table*}

\cref{tab:main_results} reports a representative operating point (means over $3$ seeds; std in the suppl.). Stale-OEM's $15.3^\circ$ absolute-rotation error is essentially the uncorrected global reorientation. \emph{Per-camera} BA, even converged, recovers poor parameters ($2.30^\circ$, $503.88$\,px focal): each single-camera solve trades focal against depth to fit reprojection ($0.16$\,px), a consistency limit, not an optimizer defect. The \emph{rig-coupled joint} BA is far stronger ($0.47^\circ$, $8.88$\,px), yet \textsc{RECALL} surpasses it on every parameter-space metric (absolute and relative rotation, translation, focal, and principal point), trailing only on the diagnostic reprojection, in a single $0.12$\,ms pass ($10^5\times$ faster than joint BA, $10^4\times$ than per-camera).
 The advantage is not a convergence artifact: the joint solve is run to the large iteration budget of \cref{subsec:baselines}. The refinement closes the gap, monotonically reducing every error over its iterations and cutting reprojection from the single-pass estimate (full per-iteration ablation in the suppl.).

\subsection{Robustness Sweep}
\label{subsec:sweep}
\vspace{-5pt}
We sweep per-camera drift $\kappa\in\{0.02,\dots,0.10\}$ (at $20^\circ$ tilt) and tilt $\in\{0^\circ,\dots,30^\circ\}$ (at $\kappa{=}0.05$), with the drift level fixed at $\kappa$; full curves in suppl. \textsc{RECALL} is more accurate than \emph{both} BA variants in rotation and focal at \emph{every} level: it degrades gracefully with drift (rotation $0.15^\circ\!\to\!0.51^\circ$ from $\kappa{=}2\%$ to $10\%$) and is nearly flat across tilt ($0.30^\circ\!\to\!0.32^\circ$), confirming robustness to reorientation. Stale-OEM grows with whatever it ignores; per-camera BA stays poor ($1.7-2.4^\circ$, $59$-$3255$\,px); the BA variants keep an edge only on the diagnostic reprojection.

\subsection{Latency}
\label{subsec:latency}
\vspace{-5pt}
Online recalibration must keep up with the capture loop. We report mean wall-clock per rig (ms; for \textsc{RECALL}, one batched GPU pass over the test set divided by the number of rigs), timing \emph{all} methods on the same NVIDIA H100 node:
\textsc{RECALL} runs at ${\sim}0.12$\,ms per rig, versus ${\sim}0.2$-$0.7$\,s for per-camera BA and tens of seconds for joint BA ($\gtrsim 10^5\times$, growing with its budget).
Times are for the recalibration operator \emph{given correspondences} (as for BA); fiducial detection is upstream.  Its cost is amortized to training, yielding constant-time inference, whereas BA pays a growing per-frame optimization cost (\cref{tab:main_results}). \textsc{RECALL}'s latency stays flat as the rig grows from $4$ to $20$ cameras while joint-BA latency scales with it, and BA's accuracy plateaus near $0.85$-$1.1^\circ$ regardless of iteration budget (only $25\%$ converged at $30$k iterations), so \textsc{RECALL} matches converged BA at ${\sim}10^5\times$ lower cost (suppl.).

\subsection{Real-Data Evaluation}
\label{subsec:realdata}
\vspace{-5pt}
We validate on TotalCapture~\cite{trumble2017total} (a fixed $8$-camera rig with independent Vicon 3D) and CMU Panoptic~\cite{joo2017panoptic} (a $10$-camera HD surround rig with reconstructed \texttt{coco19} keypoints). The published calibration is the OEM anchor; the known 3D structure is projected through the cameras with per-camera visibility (a detector sees only in-frame points). We evaluate two settings: \textbf{(i) drift-only} ($\kappa{=}0.1$), for the head-to-head against image-only foundation models, which run on the real images and so see the real, unmodified camera geometry; and \textbf{(ii) the full movable-rig task}, additionally composing a bounded $20^\circ$ global reorientation $R_G$ onto real structure (full-test-set means in suppl.). \textsc{RECALL} is trained on synthetic structures on Panoptic (a synthetic-to-real transfer) and on real markers of disjoint training subjects on TotalCapture, and evaluated on the \emph{real} structure, and we report the gauge-invariant RRA@$5^\circ$/RTA@$5^\circ$/AUC@$30^\circ$ of \cref{subsec:metrics}.

\textbf{Foundation-model comparison.} The Panoptic RGB lets us compare against the strongest image-only geometry predictors (VGGT~\cite{wang2025vggt}, VGGT-$\Omega$~\cite{vggtomega2025}, $\pi^3$~\cite{pi3_2025}) on one synchronized frame per camera. These are prior-free and recover cameras only up to a similarity; for a fair metric comparison, we lift each via an \emph{oracle} Sim(3) to the reference (their best case), so this contrasts method classes (image-only SfM vs.\ prior-anchored recalibration), not identical inputs.
\begin{table*}[tb]
    \caption{\textbf{Real-data comparison on CMU Panoptic} (drift-only; $10$-camera HD surround rig, $\kappa{=}0.1$, $94.8\%$ visibility). Image-only models (top) predict up-to-similarity; their absolute rotation and focal use an \emph{oracle} Sim(3) lift (best case), while RRA/RTA/AUC are gauge-invariant. Prior-anchored methods (bottom) receive the OEM anchor and 3D structure and are scored on a common $100$ drift draws (BA's per-instance cost prohibits the full set); foundation models are over $5$ frames. Joint BA is iteration-limited ($8\%$ converged at the $8000$-iteration budget). \textbf{Best} per metric in bold, \underline{second best} underlined; arrows ($\uparrow$/$\downarrow$) give the better direction; RRA/RTA/AUC in \%.}
    \centering
    \resizebox{0.8\textwidth}{!}{%
      \rowcolors{2}{white}{gray!18}
      \begin{tabular}{@{}ll|c|cc|c|c|c@{}}
        \toprule
        \textbf{Method} & \textbf{Input} &
        \specialcell[t]{\textbf{Abs. rot}~$\downarrow$\\(\si{\degree})} &
        \textbf{RRA@$5^\circ$}~$\uparrow$ & \textbf{RTA@$5^\circ$}~$\uparrow$ &
        \textbf{AUC@$30^\circ$}~$\uparrow$ &
        \specialcell[t]{\textbf{Focal}~$\downarrow$\\(px)} &
        \specialcell[t]{\textbf{Latency}~$\downarrow$\\(ms)} \\
        \midrule
        VGGT~\cite{wang2025vggt}                  & images      & 39.59 & 26.7 & 25.8 & 43.3 & 292.9 & 3154 \\
        VGGT-$\Omega$~\cite{vggtomega2025}        & images      & 1.16  & 99.1 & 80.0 & 89.0 & 247.2 & 2017 \\
        $\pi^3$~\cite{pi3_2025}                   & images      & 1.63  & \textbf{100.0} & 86.7 & 90.7 & 283.5 & 3196 \\
        \midrule
        Stale-OEM (do-nothing)                    & prior       & 0.72 & 100.0 & 98.2 & 96.5 & 18.78 & $\approx 0$ \\
        Per-camera BA                             & prior+corr. & 3.93 & 54.0 & 77.6 & 79.0 & 52.08 & 206 \\
        Rig-coupled joint BA ($8\%$ conv.)        & prior+corr. & \underline{0.83} & 99.87 & 98.6 & 96.7 & 17.57 & 65087 \\
        Method A (RANSAC+EPnP)                    & prior+corr. & 1.12 & 99.0 & 98.1 & 95.2 & 20.51 & \underline{3} \\
        Method B (RANSAC+EPnP)                    & prior+corr. & 1.12 & 99.0 & 98.1 & 95.2 & 18.59 & 65 \\
                Method A (REPPnP)                         & prior+corr. & 0.85 & \underline{99.89} & \underline{99.2} & \underline{96.8} & 20.51 & 11 \\
        Method B (REPPnP)                         & prior+corr. & 0.85 & \underline{99.89} & \underline{99.2} & \underline{96.8} & \underline{16.65} & 74 \\
        \textsc{RECALL} (ours)                    & prior+corr. & \textbf{0.64} & \textbf{100.0} & \textbf{99.5} & \textbf{97.7} & \textbf{16.42} & \textbf{0.18} \\

        \bottomrule
      \end{tabular}
      \label{tab:realdata}
      }
      \vspace{-10pt}
\end{table*}

\cref{tab:realdata} reports the drift-only Panoptic comparison. The foundation models are strong on \emph{relative rotation} ($\pi^3$/VGGT-$\Omega$ reach RRA@$5^\circ$ $100$/$99$ from images, not strawmen) but drop on the stricter RTA@$5^\circ$ ($87$/$80$ vs.\ ${\sim}99$ for the prior-anchored methods), recover cameras only up to scale, and estimate intrinsics poorly ($247$-$293$\,px focal); VGGT collapses on this wide-baseline surround geometry (AUC@$30^\circ$ $43.3$), and all three run at $2.0$-$3.2$\,s. \textsc{RECALL} \emph{surpasses} the joint BA on every parameter-space metric: absolute rotation ($0.64^\circ$ vs.\ $0.83^\circ$), the gauge-invariant RRA/RTA/AUC, and focal ($16.4$ vs.\ $17.6$\,px, recovered via the OEM anchor), all at $0.18$\,ms versus $65.1$\,s per rig ($\sim\!10^5\times$ faster); it also recovers focal ${\sim}15\times$ better than the foundation models and is metric. The joint BA is iteration-limited ($8\%$ converged even at $8000$ iterations); per-camera BA is poor in rotation and focal.

\noindent\textbf{Recovering the global reorientation on real data.} \cref{tab:realdata_G} adds a bounded $20^\circ$ global reorientation (the full movable-rig task, leaving the anchor $20^\circ$ off). Solving the gauge in-network (\cref{eq:gngauge}) closes the global-orientation gap a regressor would leave: \textsc{RECALL} recovers the globally reoriented rig in ${\sim}0.2$\,ms and \emph{surpasses} the offline joint BA on absolute rotation on TotalCapture ($0.75^\circ$ vs.\ $1.79^\circ$) while matching it within noise on the harder $19$-keypoint Panoptic surround rig ($0.78^\circ$ vs.\ $0.83^\circ$), leading on the relative-geometry metrics (RRA@$5^\circ$, AUC@$30^\circ$) on both, at ${\sim}10^5\times$ lower latency. The joint BA is iteration-limited ($13$-$53\%$ converged at $8000$ iterations, $34$-$61$\,s per rig) and barely improves with more iterations (suppl.). This confirms the shared global rotation is recovered to per-instance-optimization quality on real structure, not only in synthesis; BA recovers slightly more focal under the single-snapshot tilt, where focal is weakly observable.
\begin{table}[t]
    \caption{\textbf{Movable-rig recalibration on real data (global reorientation $+$ drift).} A bounded $20^\circ$ global reorientation is composed onto the real structure, leaving the stale anchor ${\approx}20^\circ$ off. Joint BA is run to an $8000$-iteration budget (convergence fraction in parentheses); \textsc{RECALL} and BA are scored on a common set of $100$ draws (\textsc{RECALL}'s full-test-set means in the suppl.). \textbf{Best} per metric in bold, \underline{second best} underlined; arrows ($\uparrow$/$\downarrow$) give the better direction.}

 \label{tab:realdata_G}
    \centering
    \setlength{\tabcolsep}{4pt}
    \resizebox{\columnwidth}{!}{%
    \begin{tabular}{@{}lccccc@{}}
        \toprule
        Method & abs.\ rot ($^\circ$)~$\downarrow$ & RRA@$5^\circ$~$\uparrow$ & AUC@$30^\circ$~$\uparrow$ & focal (px)~$\downarrow$ & lat.\ (ms)~$\downarrow$ \\
        \midrule
        \multicolumn{6}{@{}l}{\emph{TotalCapture ($8$-camera)}} \\
        Stale-OEM                    & 20.02 & 100.0 & 97.3 & 15.68 & $\approx 0$ \\
        Per-camera BA                & 5.00 & 45.6 & 74.9 & 69.33 & 219 \\
        Joint BA ($53\%$ conv.)      & 1.79 & 89.2 & 90.8 & 13.77 & 33645 \\
        Method A (RANSAC+EPnP)       & 1.04 & 98.1 & 95.9 & 14.55 & \underline{3} \\
        Method B (RANSAC+EPnP)       & 1.04 & 98.1 & 95.9 & \underline{12.97} & 62 \\
        Method A (REPPnP)            & \underline{0.91} & \underline{99.3} & \underline{96.6} & 14.55 & 10 \\
        Method B (REPPnP)            & \underline{0.91} & \underline{99.3} & \underline{96.6} & \textbf{12.68} & 69 \\
        \textsc{RECALL} (ours)       & \textbf{0.75} & \textbf{100.0} & \textbf{98.0} & 14.51 & \textbf{0.20} \\ 
        \midrule
        \multicolumn{6}{@{}l}{\emph{CMU Panoptic ($10$-camera surround)}} \\
        Stale-OEM                    & 20.03 & 100.0 & 96.5 & 18.78 & $\approx 0$ \\
        Per-camera BA                & 4.02 & 51.2 & 78.3 & 59.12 & 216 \\
        Joint BA ($13\%$ conv.)      & \underline{0.83} & 99.76 & \underline{96.6} & \underline{17.60} & 60814 \\
        Method A (RANSAC+EPnP)       & 1.14 & 98.8 & 95.1 & 20.51 & \underline{3} \\
        Method B (RANSAC+EPnP)       & 1.14 & 98.8 & 95.1 & 19.04 & 66 \\
        Method A (REPPnP)            & 0.88 & \underline{99.96}& 96.6 & 20.51 & 11 \\
        Method B (REPPnP)            & 0.88 & \underline{99.96} & 96.6 & \textbf{16.88} & 75 \\
        \textsc{RECALL} (ours)       & \textbf{0.78} & \textbf{100.0} & \textbf{97.1} & 19.75 & \textbf{0.22} \\
        \bottomrule
    \end{tabular}}
\end{table}

\vspace{-10pt}
\section{Conclusion}
\vspace{-5pt}
We presented \textsc{RECALL}, a feed-forward method for real-time recalibration of \emph{movable} multi-camera rigs that predicts, from the OEM anchor and a brief observation of a known object, a shared global rotation and small OEM-anchored per-camera drift, refined by a rig-coupled recurrent operator that \emph{solves} for the shared rotational gauge in-network with a learned Gauss-Newton step, trained entirely on on-the-fly synthetic data. Against a stale-OEM bound, per-camera BA, and a strong rig-coupled \emph{joint} BA, \textsc{RECALL} matches or beats both on absolute and relative rotation while running four to five orders of magnitude (${\sim}10^5\times$) faster. The lesson: a single per-camera view is consistency-limited and reprojection alone does not certify parameters, whereas rig-level consistency, the OEM prior, and \emph{solving} (rather than regressing) the shared rotational gauge recover them.

\noindent\textbf{Limitations and future work.} A model is trained per rig/object configuration; the learned gauge solver extrapolates well beyond the trained tilt range (supplementary), but very large reorientations (free axial rotation, near-$90^\circ$ tilt) remain outside the bounded-rotation assumption, and gross correspondence outliers degrade accuracy absent a robust front-end. Focal length is only weakly observable from a single snapshot, so the observability-gated loss defers to the OEM prior where it cannot be recovered, and multiple snapshots sharpen it. Promising directions include cross-rig generalization without retraining, an outlier-robust front-end, and a calibrated uncertainty estimate that flags when a rig needs recalibration.

\section*{Acknowledgment}
\label{sec:ack}

This research was undertaken, in part, based on support from the Natural Sciences and Engineering Research Council of Canada Grant RGPIN-2026-06934 (NSERC DG) and the Mexican Agency for International Development Cooperation (AMEXCID) through the Quebec Resident-Rate Tuition Program (Programa de Colegiaturas Tipo Residente del Gobierno de la Provincia de Quebec).
{
    \small
    \bibliographystyle{ieeenat_fullname}
    \bibliography{egbib}
}

\end{document}